\documentclass[11pt,letterpaper]{article}
\usepackage{acl2017}
\usepackage{times}
\usepackage{latexsym}
\usepackage{pbox}
\usepackage[acronym]{glossaries}
\usepackage{amssymb}
\usepackage{lipsum}
\usepackage{graphicx}
\usepackage[colorinlistoftodos,prependcaption,textsize=tiny]{todonotes}
\usepackage{multirow}

\aclfinalcopy

\newcommand{\EntitySpace}{\mathcal{E}}                   
\newcommand{\RelationSpace}{\mathcal{R}}

\newcommand{\head}{\mathbf{h}}
\newcommand{\rel}{\mathbf{r}}
\newcommand{\tail}{\mathbf{t}}

\newacronym{kbc}{KBC}{knowledge base completion}
\newacronym{nll}{NLL}{negative log-likelihood}
\newacronym{mrr}{MRR}{mean reciprocal rank}
\newacronym{mr}{MR}{mean rank}

\newcommand{\dg}{\textsuperscript{\dag}}
\newcommand{\ddg}{\textsuperscript{\ddag}}
\newcommand{\dddg}{$^\sharp$}

\title{Knowledge Base Completion: Baselines Strike Back}

\author{Rudolf Kadlec \and Ondrej Bajgar \and Jan Kleindienst \\
  IBM Watson\\
  V Parku 4, 140 00 Prague, Czech Republic\\
  {\tt \{rudolf\_kadlec, obajgar, jankle\}@cz.ibm.com }}

\date{}

\begin{document}

\maketitle

\begin{abstract}

Many papers have been published on the knowledge base completion task in the past few years. Most of these introduce novel architectures for relation learning that are evaluated on standard datasets such as FB15k and WN18. This paper shows that the accuracy of almost all models published on the FB15k can be outperformed by an appropriately tuned baseline --- our reimplementation of the DistMult model. 
Our findings cast doubt on the claim that the performance improvements of recent models are due to architectural changes as opposed to hyper-parameter tuning or different training objectives.
This should prompt future research to re-consider how the performance of models is evaluated and reported.


\end{abstract}

\section{Introduction}


Projects such as Wikidata\footnote{https://www.wikidata.org/} or earlier Freebase~\cite{Bollacker:2008:FCC:1376616.1376746} have successfully accumulated a formidable amount of knowledge in the form of $\langle\text{entity1 - relation - entity2}\rangle$ triplets. Given this vast body of knowledge, it would be extremely useful to teach machines to reason over such knowledge bases. One possible way to test such reasoning is \gls{kbc}.

The goal of the \gls{kbc} task is to fill in the missing piece of information into an incomplete triple. For instance, given a query $\langle \text{Donald Trump, president of, ?}\rangle$ one should predict that the target entity is USA.

More formally, given a set of entities $\EntitySpace$ and a set of binary relations $\RelationSpace$ over these entities, a \emph{knowledge base} (sometimes also referred to as a knowledge \emph{graph}) can be specified by a set of triplets $\langle h,r,t \rangle$ where $h,t \in \EntitySpace$ are head and tail entities respectively and $r \in \RelationSpace$ is a relation between them. In \emph{entity} \gls{kbc} the task is to predict either the tail entity given a query $\langle h,r,? \rangle$, or to predict the head entity given $\langle ?,r,t \rangle$.

Not only can this task be useful to test the generic ability of a system to reason over a knowledge base, but it can also find use in expanding existing incomplete knowledge bases by deducing new entries from existing ones.


An extensive amount of work has been published on this task (for a review see \cite{Nickel2015,Nguyen2017}, for a plain list of citations see Table~\ref{tab:results}). Among those DistMult~\cite{Yang2015} is one of the simplest.\footnote{We could even say too simple given that it assumes symmetry of all relations which is clearly unrealistic.}
Still this paper shows that even a simple model with proper hyper-parameters and training objective evaluated using the standard metric of Hits@10 can outperform 27 out of 29 models which were evaluated on two standard \gls{kbc} datasets,  WN18 and FB15k~\cite{Bordes2013}.

This suggests that there may be a huge space for improvement in hyper-parameter tuning even for the more complex models, which may be in many ways better suited for relational learning, e.g. can capture directed relations.



\section{The Model}

Inspired by the success of word embeddings in natural language processing, distributional models for \gls{kbc} have recently been extensively studied. Distributional models represent the entities and sometimes even the relations as $N$-dimensional real vectors\footnote{Some models represent relations as matrices instead.}, we will denote these vectors by bold font, $\head, \rel, \tail \in \mathbb{R}^N$. 

The DistMult model was introduced by \citet{Yang2015}. Subsequently \citet{Toutanova2015b} achieved better empirical results with the same model by changing hyper-parameters of the training procedure and by using negative-log likelihood of softmax instead of L1-based max-margin ranking loss. \citet{Trouillon2016} obtained even better empirical result on the FB15k dataset just by changing DistMult's hyper-parameters.


DistMult model computes a score for each triplet $\langle \head, \rel, \tail \rangle$ as
$$
s( \head, \rel, \tail) = \head^T \cdot W_\rel \cdot \tail = \sum_{i=1}^{N}{h_i r_i t_i}
$$

where $W_\rel$ is a diagonal matrix with elements of vector $\rel$ on its diagonal. Therefore the model can be alternatively rewritten as shown in the second equality. 

In the end our implementation normalizes the scores by a softmax function. That is
$$
P(t|h,r) = \frac{exp(s(h,r,t))}{\sum_{\bar{t} \in \EntitySpace_{h,r} }{exp(s(h,r,\bar{t}))}}
$$
where $\EntitySpace_{h,r}$ is a set of candidate answer entities for the $\langle h,r,? \rangle$ query.  

\section{Experiments}
\textbf{Datasets.}
In our experiments we use two standard datasets  WN18 derived from WordNet~\cite{fellbaum1998wordnet} and FB15k derived from the Freebase knowledge graph~\cite{Bollacker:2008:FCC:1376616.1376746}. 


\textbf{Method.}
For evaluation, we use the \emph{filtered} evaluation protocol proposed by \citet{Bordes2013}. 
During training and validation we transform each triplet $\langle h,r,t \rangle$ into two examples: tail query $\langle h,r,? \rangle$ and head query $\langle ?,r,t \rangle$. We train the model by minimizing \gls{nll} of the ground truth triplet $\langle h,r,t \rangle$ against randomly sampled pool of $M$ negative triplets $\langle h,r,t' \rangle, t' \in \EntitySpace \setminus \{t\}$ (this applies for tail queries, head queries are handled analogically).

In the filtered protocol we rank the validation or test set triplet against all corrupted (supposedly untrue) triplets -- those that do not appear in the train, valid and test dataset (excluding the test set triplet in question itself). 
Formally, for a query $\langle h,r,? \rangle$ where the correct answer is $t$, we compute the rank of $\langle h,r,t \rangle$ in a candidate set $C_{h,r} = \{\langle h,r,t' \rangle\ : \forall t' \in \EntitySpace \} \setminus (Train \cup Valid \cup Test) \cup \{\langle h,r,t \rangle\}$, where $Train$, $Valid$ and $Test$ are sets of true triplets.
Head queries $\langle ?,r,t \rangle$ are handled analogically.
Note that softmax normalization is suitable under the filtered protocol since exactly one correct triplet is guaranteed to be among the candidates.

In our preliminary experiments on FB15k, we varied the batch size $b$, embedding dimensionality $N$, number of negative samples in training $M$, L2 regularization parameter and learning rate $lr$. Based on these experiments we fixed lr=0.001, L2=0.0 and we decided to focus on influence of batch size, embedding dimension and number of negative samples. For final experiments we trained several models from hyper-parameter range: $N \in \{128,256,512,1024\}$, $b \in \{16,32,64,128,256,512,1024,2048\}$ and $M \in \{20, 50, 200, 500, 1000, 2000\}$. 

We train the final models using Adam~\cite{Kingma2015} optimizer ($lr=0.001,\beta_1=0.9, \beta_2=0.999, \epsilon=10^{-8}, decay=0.0$).  We also performed limited experiments with Adagrad, Adadelta and plain SGD. Adagrad usually required substantially more iterations than ADAM to achieve the same performance.
We failed to obtain competitive performance with Adadelta and plain SGD. 
On FB15k and WN18 validation datasets the best hyper-parameter combinations were $N=512$, $b=2048$, $M=2000$ and $N=256$, $b=1024$, $M=1000$, respectively. Note that we tried substantially more hyper-parameter combinations on FB15k than on WN18.
Unlike most previous works we do not normalize neither entity nor relation embeddings.

To prevent over-fitting, we stop training once Hits@10 stop improving on the validation set. 
On the FB15k dataset our Keras~\cite{Chollet2015} based implementation with TensorFlow~\cite{Abadi2015} backend needed about 4 hours to converge when run on  a single GeForce GTX 1080 GPU.


\textbf{Results.}
Besides single models, we also evaluated performance of a simple ensemble that averages predictions of multiple models. This technique consistently improves performance of machine learning models in many domains and it slightly improved results also in this case. 

The results of our experiments together with previous results from the literature are shown in Table~\ref{tab:results}. 
DistMult with proper hyperparameters twice achieves the second best score and once the third best score in three out of four commonly reported benchmarks (\gls{mr} and Hits@10 on WN18 and FB15k). On FB15k only the IRN model~\cite{shen2016implicit} shows better Hits@10 and the ProjE~\cite{Shi2017} has better \gls{mr}. 

Our implementation has the best reported \gls{mrr} on FB15k, however this metric is not reported that often. \gls{mrr} is a metric of ranking quality that is less sensitive to outliers than \gls{mr}.

On WN18 dataset again the IRN model together with R-GCN+ shows better Hits@10. However, in \gls{mr} and \gls{mrr} DistMult performs poorly. Even though DistMult's inability to model asymmetric relations still allows it to achieve competitive results in Hits@10 the other metrics clearly show its limitations. These results highlight qualitative differences between FB15k and WN18 datasets. 

Interestingly on FB15k recently published models (including our baseline) that use only $\mathbf{r}$ and $\mathbf{h}$ or $\mathbf{t}$ as their input outperform models that utilize richer features such as text or knowledge base path information. This shows a possible gap for future improvement.

Table~\ref{tab:acc} shows accuracy (Hits@1) of several models that reported this metric. On WN18 our implementation performs worse than HolE and ComplEx models (that are equivalent as shown by \citet{Hayashi2017}). On FB15k our implementation outperforms all other models.       

\subsection{Hyper-parameter influence on FB15k}
In our experiments on FB15k we found that increasing the number of negative examples $M$ had a positive effect on performance. 

\begin{figure}[!ht]
    \includegraphics[scale=0.35]{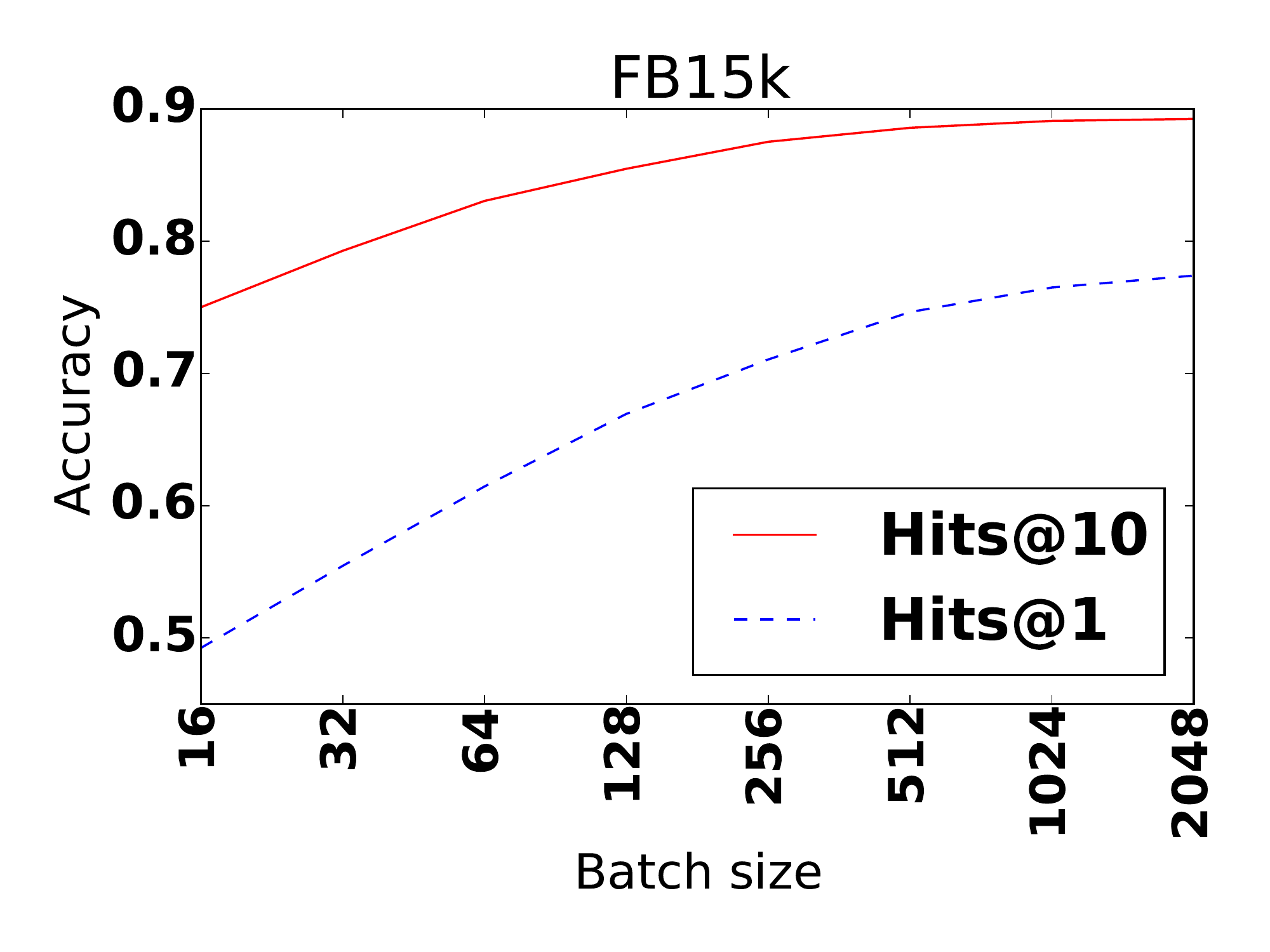}
    \caption{Influence of batch size on Hits@10 and Hits@1 metrics for a single model with $N=512$ and $M=2000$.}
    \label{fig:batch_acc_fb15k} 
\end{figure}

Another interesting observation is that batch size has a strong influence on final performance. 
Larger batch size always lead to better results, for instance  Hits@10 improved by $14.2$\% absolute when the batch size was increased from 16 to 2048. See Figure~\ref{fig:batch_acc_fb15k} for details. 



Compared to previous works that trained DistMult on these datasets (for results see bottom of Table~\ref{tab:results}) we use different training objective than \citet{Yang2015} and \citet{Trouillon2017} that optimized max margin objective and \gls{nll} of softplus activation function ($softplus(x)=ln(1+e^x)$), respectively. Similarly to \citet{Toutanova2015b} we use \gls{nll} of softmax function, however we use ADAM optimizer instead of RProp~\cite{riedmiller1993direct}.


\begin{table}[ht]
    \centering
    \begin{tabular}{l|ll}
        \hline

          \multirow{2}{*}{ \bf Method } & \multicolumn{2}{c}{\textbf{Accuracy(Hits@1)}} \\ \cline{2-3}

       &  WN18 & FB15k \\ \hline

        HolE~\dg &  93.0 & 40.2\\ 
        DistMult~\ddg & 72.8 & 54.6  \\
        ComplEx~\ddg &  \textbf{93.6} & 59.9  \\
        R-GCN+~\dddg & 67.9 & 60.1 \\
        \hline
        \bf DistMult ensemble & 78.4 & \textbf{79.7} 

        
    \end{tabular}
    \caption{ Accuracy (Hits@1) results sorted by performance on FB15k. Results marked by \dg, ~\ddg{} and ~\dddg{} are from \cite{Nickel2016}, \cite{Trouillon2017} and \cite{Schlichtkrull2017}, respectively. Our implementation is listed in the last row.}
    \label{tab:acc}
\end{table}


\begin{table*}
\centering
\resizebox{16cm}{!}{
\begin{tabular}{l|lll|lll|l}
\hline
\multirow{3}{*}{\bf Method}& \multicolumn{6}{|c|}{\textbf{Filtered}} & \parbox[t]{2mm}{\multirow{3}{*}{\rotatebox[origin=c]{90}{\pbox{4cm}{\bf Extra \\ features}}}} \\
\cline{2-7}
& \multicolumn{3}{|c}{\bf WN18} & \multicolumn{3}{|c|}{\bf FB15k}\\
\cline{2-7}
\cline{2-7}
& MR & H10 & MRR &   MR & H10 & MRR \\
\hline
SE \citep{Bordes2011} &  985 &80.5 & - &  162  & 39.8& - & \parbox[t]{2mm}{\multirow{17}{*}{\rotatebox[origin=c]{90}{None}}} \\
Unstructured \citep{Bordes2014}  &   304 & 38.2  &-&   979 &  6.3& - \\
TransE \citep{Bordes2013}  &  251 & 89.2 & -  &  125 & 47.1 & - \\
TransH \citep{Wang2014a}  &   303 & 86.7 & - &  87 &  64.4&- \\
TransR \citep{Lin2015a}  &   225 &92.0 & - &   77 &   68.7&- \\
CTransR \citep{Lin2015a}  &  218 &  92.3& -  &   75  & 70.2 &  -\\
KG2E \citep{He2015a}  &   331 & 92.8 & - &  59 & 74.0& -  \\
TransD \citep{Ji2015}  &   212 &  92.2 & - &   91 & {77.3}&  -  \\
lppTransD \citep{Yoon2016} &  270 & 94.3 & - & 78  & 78.7 & -\\
TranSparse \citep{Ji2016} &  211  & 93.2 & - & 82  & 79.5& -\\

TATEC \citep{Garcia-Duran2016}  &  - &  - & - &   58 & 76.7  & -  \\
NTN \citep{Socher2013}  &  - & {66.1}  & 0.53& - & 41.4& 0.25  \\
HolE \citep{Nickel2016} &  -& 94.9 & \textbf{0.938} & - & 73.9 & 0.524\\ 
STransE \cite{Nguyen2016a}   &   \textbf{206}  & {93.4}& 0.657 & 69  & 79.7 & 0.543\\
ComplEx \cite{Trouillon2017} & - & 94.7 & \large{\underline{\textbf{0.941}}} & - & 84.0 & 0.692 \\
ProjE wlistwise \cite{Shi2017} & - & - & - & \large{\underline{\textbf{34}}} & 88.4 & - \\ 
IRN \cite{shen2016implicit} & 249 & \textbf{95.3} & - & \textbf{38} & \large{\underline{\textbf{92.7}}} & - \\


\hline
\hline
\textsc{r}TransE \citep{Garcia2015}  &   -& - & - & 50 & 76.2  & - \\
PTransE \citep{Lin2015}  &  - & -  & - & {58} & {84.6}& - & \parbox[t]{2mm}{\multirow{3}{*}{\rotatebox[origin=c]{90}{Path}}} \\

GAKE \citep{Feng2016} &  - & - & - & 119 & 64.8& - \\ 
Gaifman \citep{Niepert2016} &  352  & 93.9 & -& 75 & 84.2 & - \\ 
Hiri \citep{Liu2016} &  -  & 90.8 & 0.691& -  & 70.3& 0.603 \\
R-GCN+ \cite{Schlichtkrull2017} & - & \large{\underline{\textbf{96.4}}} & 0.819 & - & 84.2 & 0.696 \\

\hline
\hline

NLFeat \citep{Toutanova2015b}  &  -   & 94.3 & \textbf{0.940}& -  & 87.0& \textbf{0.822} & \parbox[t]{2mm}{\multirow{3}{*}{\rotatebox[origin=c]{90}{Text}}} \\
TEKE\_H  \citep{Wang2016} &   \large{\underline{\textbf{114}}}  & 92.9 & -& 108 & 73.0& -  \\ 
SSP \citep{Xiao2017} &  \textbf{156} & 93.2 & - & 82 & 79.0 & -\\ 

\hline
\hline
DistMult (orig) \citep{Yang2015}  &  - & {{94.2}} & 0.83 & -  & 57.7 & 0.35 & \parbox[t]{2mm}{\multirow{5}{*}{\rotatebox[origin=c]{90}{None}}}\\
DistMult \citep{Toutanova2015b}  &  - & - & - & -  & 79.7 & 0.555\\
DistMult \citep{Trouillon2017}  &  - & 93.6 & 0.822 & -  & 82.4 & 0.654\\

\cline{1-7}

\bf Single DistMult (this work) & 655 & 94.6 & 0.797 & 42.2 & \textbf{89.3} & \textbf{0.798} \\
\bf Ensemble DistMult (this work) & 457 & \textbf{95.0} & 0.790 & \textbf{35.9} & \textbf{90.4} & \large{\underline{\textbf{0.837}}} \\
\end{tabular}
}
\caption{Entity prediction results. MR, H10 and MRR denote evaluation metrics of mean rank, Hits@10 (in \%) and mean reciprocal rank, respectively. The three best results for each metric are in bold. Additionally the best result is underlined.
The first group (above the first double line) lists models that were trained only on the knowledge base and they do not use any additional input besides the source entity and the relation.
The second group shows models that use path information, e.g. they consider paths between source and target entities as additional features. The models from the third group were trained with additional textual data. In the last group we list various implementations of the DistMult model including our implementation on the last two lines. Since DistMult does not use any additional features these results should be compared to the models from the first group. 
``NLFeat'' abbreviates Node+LinkFeat model from \citep{Toutanova2015b}. The results for NTN {\protect\citep{Socher2013}} listed in this table are taken from {\protect\citet{Yang2015}}. This table was adapted from~\cite{Nguyen2017}.}
\label{tab:results}
\end{table*}

\section{Conclusion}

Simple conclusions from our work are:
1) Increasing batch size dramatically improves performance of DistMult, which raises a question whether other models would also significantly benefit from similar hyper-parameter tuning or different training objectives;
2) In the future it might be better to focus more on metrics less frequently used in this domain, like Hits@1 (accuracy) and MRR since for instance on WN18 many models achieve similar, very high Hits@10, however even models that are competitive in Hits@10 underperform in Hits@1, which is the case of our DistMult implementation.

A lot of research focus has recently been centred on the filtered scenario which is why we decided to use it in this study. An advantage is that it is easy to evaluate. However the scenario trains the model to expect that there is only a single correct answer among the candidates which is unrealistic in the context of knowledge bases. Hence future research could focus more on the \emph{raw} scenario which however requires using other information retrieval metrics such as \emph{mean average precision} (MAP), previously used in KBC for instance by \citet{Das2016}.



We see this preliminary work as a small contribution to the ongoing discussion in the machine learning community about the current strong focus on state-of-the-art empirical results when it might be sometimes questionable whether they were achieved due to a better model/algorithm or just by more extensive hyper-parameter search. For broader discussion see \cite{CHURCH2017}.

In light of these results we think that the field would benefit from a large-scale empirical comparative study of different \gls{kbc} algorithms, similar to a recent study of word embedding models~\cite{Levy2015}. 



\bibliography{references}
\bibliographystyle{emnlp_natbib}

\end{document}